\documentclass[conference]{IEEEtran}
\IEEEoverridecommandlockouts

\ifCLASSOPTIONcompsoc
\usepackage[caption=false,font=normalsize,labelfon
t=sf,textfont=sf]{subfig}
\else
\usepackage[caption=false,font=footnotesize]{subfig}
\fi

\usepackage[flushleft]{threeparttable}
\usepackage{cite}
\usepackage{amsmath,amssymb,amsfonts}
\usepackage{algorithmic}
\usepackage{graphicx}
\usepackage{textcomp}
\usepackage{xcolor}
\usepackage{balance}

\def\BibTeX{{\rm B\kern-.05em{\sc i\kern-.025em b}\kern-.08em
    T\kern-.1667em\lower.7ex\hbox{E}\kern-.125emX}}

\usepackage{tikz}
\usepackage{textcomp}
\usepackage{hyperref}
\usepackage{lipsum}

\newcommand\copyrighttext{%
  \footnotesize \textcopyright 2020 IEEE.Personal use of this material is permitted.  Permission from IEEE must be obtained for all other uses, in any current or future media, including reprinting/republishing this material for advertising or promotional purposes, creating new collective works, for resale or redistribution to servers or lists, or reuse of any copyrighted component of this work in other works.  DOI: \href{https://ieeexplore.ieee.org/document/9357187}{10.1109/CiSt49399.2021.9357187}}
\newcommand\copyrightnotice{%
\begin{tikzpicture}[remember picture,overlay]
\node[anchor=south,yshift=10pt] at (current page.south) {\fbox{\parbox{\dimexpr\textwidth-\fboxsep-\fboxrule\relax}{\copyrighttext}}};
\end{tikzpicture}%
}

\begin{document}

\title{Recognizing semantic relation in sentence pairs using Tree-RNNs and Typed dependencies}

\author{\IEEEauthorblockN{Jeena Kleenankandy}
\IEEEauthorblockA{\textit{Department of Computer Science and Engineering} \\
\textit{National  Institute of Technology Calicut}\\
Kerala, India \\
jeenakk@gmail.com}
\and
\IEEEauthorblockN{K A Abdul Nazeer}
\IEEEauthorblockA{\textit{Department of Computer Science and Engineering} \\
\textit{National  Institute of Technology Calicut}\\
Kerala, India \\
nazeer@nitc.ac.in}
}
\maketitle
\copyrightnotice
\begin{abstract}
Recursive neural networks (Tree-RNNs) based on dependency trees are ubiquitous in modeling sentence meanings as they effectively capture semantic relationships between non-neighborhood words. However, recognizing semantically dissimilar sentences with the same words and syntax is still a challenge to Tree-RNNs. This work proposes an improvement to Dependency Tree-RNN (DT-RNN) using the grammatical relationship type identified in the dependency parse. Our experiments on semantic relatedness scoring (SRS) and recognizing textual entailment (RTE) in sentence pairs using SICK (Sentence Involving Compositional Knowledge) dataset show encouraging results. The model achieved a 2\% improvement in classification accuracy for the RTE task over the DT-RNN model. The results show that Pearson's and Spearman's correlation measures between the model's predicted similarity scores and human ratings are higher than those of standard DT-RNNs.  
\end{abstract}


\section{Introduction}
Measuring Semantic Similarity between sentence pairs is crucial for many Natural Language Processing (NLP) tasks, including but not limited to Sentence Classification \cite{liu2019bidirectional,xia2018novel}, Natural Language Inference \cite{choi2018learning,xiong2020dgi}, Paraphrase Identification \cite{agarwal2018deep,jang2019recurrent}, Question Answering \cite{liu2019visual,zhu2020knowledge}, Sentiment Analysis \cite{kim-2014-convolutional,tai2015improved}, and Semantic Relatedness Scoring \cite{shen2020learning,tien2019sentence}. The semantics of a text is a composition of the meaning of its words and syntax. Recursive neural networks (RecNN) and its variants have become ubiquitous for semantic composition as they reflect the language's inherent tree structure: word form phrases, phrases form larger phrases or sentences. A neural network at each node of RecNN computes the node's vector representation recursively from the vectors of its child nodes. RecNNs are hence also known as Tree-RNNs. The root node's vector denotes the sentence embedding. These recursive NN are either modeled on dependency parse trees, constituency parse trees \cite{socher2010learning,socher2011parsing,socher2012semantic,socher2013recursive,tai2015improved}, or latent trees learned for specific tasks\cite{choi2018learning,yogatama2016learning}.
\begin{figure*}[!t]
\centering
\subfloat[the fish is following the turtle]{\includegraphics[width=2.5in]{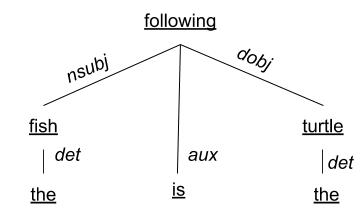}
}
\hfil
\subfloat[the turtle is following the fish]{\includegraphics[width=2.5in]{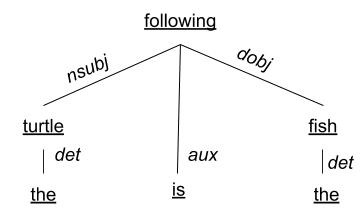}
}
\caption{An example of two phrases that have high word and syntactic overlap but are semantically contradicting. These dependency trees are mirror images of each other except for their edge labels.}
\label{fig:turtlefish}
\end{figure*}
Dependency parse structures are more apt for modeling textual meaning as they reflect the semantic relationship between words in the sentence. As standard RecNN\cite{socher2014tacl} models based on dependency parse focus only on the words and the tree structure, they fail to differentiate between sentences with the same words and syntax but with different meanings. For example, the dependency parse trees of the two phrases ``the fish followed the turtle" and ``the turtle followed the fish" has the same structure and nodes (Fig. \ref{fig:turtlefish}). The difference lies in their edge labels, which are grammatical relation between word pairs, also known as Typed dependencies\cite{de2008stanford,schuster2016enhanced}. In the first sentence, the dependency relationships of the pairs (following, fish) and (following, turtle) are nominal subject (nsubj) and direct object (dobj), respectively. These relations are reversed in the second sentence. A Tree-RNN that does not take the edge labels as input cannot semantically differentiate these sentences. 

This research aims to design a recursive neural network over the dependency parse tree that can include both the nodes and the edge labels to learn better sentence vectors and thereby compute the semantic relation between them.  Our proposed Typed DT-RNN model composes sentence embeddings using a separate dependency vector for each grammatical relation in the tree. Using a siamese architecture, the model learns sentence vectors for each sentence pair and uses it to predict the relatedness score and textual entailment between the sentences in it. We use the standard SICK (Sentence Involving Compositional Knowledge) \cite{marelliraffaella} dataset for experiments. The proposed model's accuracy improved by $2\%$ using typed dependencies in dependency tree RNNs for entailment classification. The semantic relatedness scores predicted by the model show a higher correlation with human ratings than other recursive NN models. \\
The main contributions of this paper are :
\begin{enumerate}
 \item We propose an enhancement to DT-RNN by using dependency types to learn better sentence semantic representation.
\item We present the experimental results of the proposed model on two NLP tasks: Semantic relatedness scoring, and Recognizing Textual Entailment, using the standard SICK dataset.
\end{enumerate}

The rest of this paper is organized as follows. Section \ref{sec:background} explains the standard recursive neural networks over dependency parse trees used for sentence semantic modeling. A brief survey of related works is given in section \ref{sec:related}. Section \ref{sec:method} details the sentence encoding and classification framework used in this work. The proposed Typed DT-RNN is explained in section \ref{sec:tydtrnn}. The experimental settings and its results are discussed in sections \ref{sec:exp} and \ref{sec:results} respectively. We conclude the paper in section \ref{sec:Conclusion}.

\section{Background: Dependency Tree - RNN}\label{sec:background}
This section explains how standard recursive neural networks compose sentence vectors, from its word embeddings and dependency tree, with two popular models: DT-RNN and SDT-RNN. 
Recursive neural networks or Tree-RNN are deep learning networks used for modeling tree-structured data and hence are most appropriate for natural language text. Tree-RNN repeatedly applies a transition function on the tree nodes to compute the node's vector representation from its child nodes. As the number of child nodes of a dependency tree is not fixed, the transition function at every parent node takes the summation of the output vectors of child nodes for composition. Equation \ref{eqn:dtrnn} denotes the transition function of DT-RNN at node $t$ whose child nodes are denoted by the set $C(t)$.
\begin{equation}
\label{eqn:dtrnn}
h_t = f \left( \frac{1}{l(t)} \left(W_v x_t + \sum_{k \epsilon C(t)} l(k) W_{loc_{tk}} h_k \right) \right)
\end{equation}
where $x_t$ is the vector of the word at t, $W_v \epsilon R^{h \times d}$ and $W_{loc_{tk}} \epsilon R^{h \times h}$ are the weight matrices of linear layers where the subscript $loc_{tk}$ refers to the relative location of $k$ from $t$, $l(t)$ is the number of leaf nodes(words) under the node $t$, $h_k$ is the hidden state of the subtree rooted at k, and $f$ is a non-linear activation function. The size of hidden state $h$ is a hyperparameter of the model and $d$ is the dimension of word embedding.  For a word with two dependents on its left and one on its right, the weight matrices shall be $W_{l1}$, $W_{l2}$, and $W_{r1}$, respectively. The number of weight matrices required depends on the length of the sentences in the training set. The summation term is absent in leaf nodes. 
In Semantic DT-RNN or SDT-RNN, the  $W_{loc_{tk}}$ matrices in equation \ref{eqn:dtrnn} is replaced by  $W_{dep_{tk}}$, where $dep_{tk}$ is the type of dependency relation between the parent node $t$ and its child node $k$.  In \cite{tai2015improved}, the authors report that DT-RNN performs slightly better than SDT-RNN for semantic relatedness scoring on the SICK dataset. 

\section{Related works}\label{sec:related}
Training a single shared neural network to compose words of different characteristics and with different grammatical relations has always been a challenge for recursive neural networks.  Many works in the literature attempt to overcome this hurdle using multiple weight matrices. For instance, Matrix-Vector RNN (MV-RNN)\cite{socher2012semantic} learns different weight matrices for every word in the vocabulary, making it highly complex to train. In Syntactically untied RNN (SU-RNN) \cite{socher2013parsing}, the function to compute a parent vector depends on the syntactic categories of its children. On the other hand, Tag-guided RNN (TG-RNN) \cite{qian2015learning} chooses its composition function based on the syntactic tag of the parent node.  The Tag Embedded RNN (TE-RNN) \cite{qian2015learning} learns an embedding of the syntactic tags of the left and right child nodes, and its function takes as input these tag vectors concatenated with the phrase vectors for composition. Adaptive Multi-Compositionality (AdaMC) RNN \cite{dong2014adaptive} eliminates the need for tag information by adapting its composition function to a weighted combination of a fixed number of functions depending on the input vectors. Tag-Guided HyperRecNN \cite{shen2020learning} learns a dynamic semantic composition function with the help of a recursive hyper network of Part-of-Speech (POS) tags. 
These Tree-RNN models use constituency trees for sentence classification and do not scale up for dependency trees. Unlike constituency trees, which are binarised and have words only at its leaves, dependency trees are n-ary trees with words at every node.
Socher et al.'s Dependency Tree-RNN (DT-RNN) and Semantic DT-RNN (SDT-RNN)  \cite{socher2014tacl} are the pioneering works in deep learning that uses dependency trees for semantic modeling. These models handle the problem of single shared composition function using different weight matrices for each child node depending on its relative position or edge label in the tree. Both these models have a large number of parameters making it computationally expensive to train and easy to overfit to the training data.
The idea of using grammatical roles for improving semantic modelling has been proposed in many recent sequential\cite{10.1371/journal.pone.0193919,zhu2019part} as well as recursive\cite{tai2015improved,kim2019dynamic,shen2020learning,wang2017tag,10.1371/journal.pone.0193919,zhu2019part,kleenankandy2020enhanced} Long Short Term Memory (LSTM) models. LSTM are a type of recursive NN that uses additional cell state along with gating vectors to select and retain information over longer sequences. They perform better than standard RNNs but with additional computational and memory costs. The focus of this work is on improving standard Tree-RNNs.

\section{Method}\label{sec:method}
\begin{figure}
  \centering
  \includegraphics[width=0.85\linewidth]{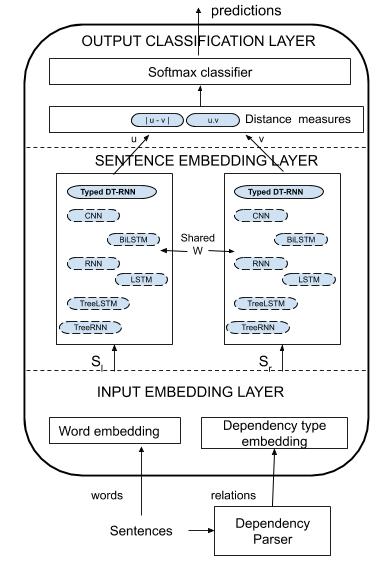}
  \caption{A general framework widely adopted for sentence pair modeling.}
  \label{fig:framework}
\end{figure}

This section briefly explains the sentence pair modeling approach\cite{tai2015improved} used in this work. Fig. \ref{fig:framework} shows the general framework of a training scheme widely used for semantic matching tasks. The input embedding layer encodes the input words and grammatical information into dense vectors. The sentence embedding layer composes semantic vector pair for each pair of sentences using a siamese network with tied weights. The output classification layer computes the vector differences to classify the sentence relations using a softmax classifier. The figure shows the proposed Typed DT-RNN network in the sentence embedding layer. Other alternative models are also shown in the figure.

The input to this experiment is a set of triplets ($S_l, S_r, y$). For semantic relatedness scoring $y \,( 1 \leq y \geq K$) is the degree of semantic similarity between the sentences $S_l$ and $S_r$. We convert $y$ into a probability distribution $p$ over $K$ classes using the equation \ref{eqn:srsp}\cite{tai2015improved}.  The value of $K=5$ corresponds to the highest similairty score.  For $1 \leq i \leq K$,
\begin{equation}
\label{eqn:srsp}
p_i = \begin{cases} y - \lfloor y \rfloor, & i = \lfloor y \rfloor + 1 \\ \lfloor y \rfloor - y + 1, & i = \lfloor y \rfloor  \\ 0 & \text{otherwise}\end{cases}
\end{equation}

For classifying entailment relation, the actual class label $y$ is converted to a probability vector $p$ with the value $1$ for the correct label and $0$ elsewhere. The dimension of probability vector  $K = 3$ corresponding to the classes -- \emph{Contradiction, Neutral, and Entailment}. The goal of the network is to minimize the KL-divergence between the actual probability and the predicted one.

The sentence encoding process is as follows: 
Given a sentence or phrase $S$ as an ordered list of n-words ($w_1,w_2, ..., w_n$), the first step is to parse $S$ to get its dependency tree. The dependency tree is a rooted tree whose nodes are the words of the sentences, and each edge label represents the grammatical relation between the words it connects. 
Next, we map each word in the sentence to a $d$-dimensional vector to embed its meaning.  The typed dependencies are also embedded into $r$-dimensional vectors called as typed dependency embeddings\cite{kleenankandy2020enhanced}. The input to the Typed DT-RNN for each sentence is a set of word vectors, typed dependency embeddings, and its dependency tree. The Typed DT-RNN recursively computes each node's vector representation from its child nodes using the word vectors and the dependency embeddings. The vector at the root represents the sentence vector.  As shown in Fig. \ref{fig:tydtrnn}, a pair of vectors $u$ and $v$ are generated for each sentence pair using a siamese architecture with tied weights. The two vector's absolute difference and dot product are input to a neural network classifier (Equation \ref{eqn:srs}). A softmax layer predicts the probability of the sentence pair over $K$ classes.

For relatedness scoring, the predicted probability vector $\hat{p}$ is then converted to the predicted similarity score $\hat{y}$.
\begin{align}
 h_s &= \tanh \left( W_{c}[(u \odot v) : |u - v|] + b_c \right), W_c \, \epsilon \, R^{2h\times c}\nonumber\\\label{eqn:srs} 
 \hat{p}_\theta &= softmax(W_{p}h_s + b_p), W_{p} \, \epsilon \, R^{c \times K}\\\nonumber
 \hat{y} &=  r^T \hat{p}_\theta\; ,\; r = [1,2,3...,K]
\end{align}

In equation \ref{eqn:srs}, $c$ the size of hidden layer of classifier is hyperparameter to be optimized.
The model is trained by back-propagation to minimize the regularized KL-divergence between $p^{(j)}$ and $\hat{p}^{(j)}_\theta$ for each sentence pair $(S_l, S_r)^{(j)}$ in the training set of size $m$.
\begin{equation}
J(\theta) = \frac{1}{m} \displaystyle\sum_{j=1}^m KL \left(p^{(j)}\|\hat{p}^{(j)}_\theta \right) + \frac{\lambda}{2}\|\theta\|^2_2,
\end{equation}

\section{Typed Dependency Tree RNN}\label{sec:tydtrnn}

Our proposed model is an enhancement to DT-RNN that encodes sentences into fixed dimensional vectors using the word embedding, dependency tree structure, and dependency relation between words.
\begin{figure}
  \centering
  \includegraphics[width=\linewidth]{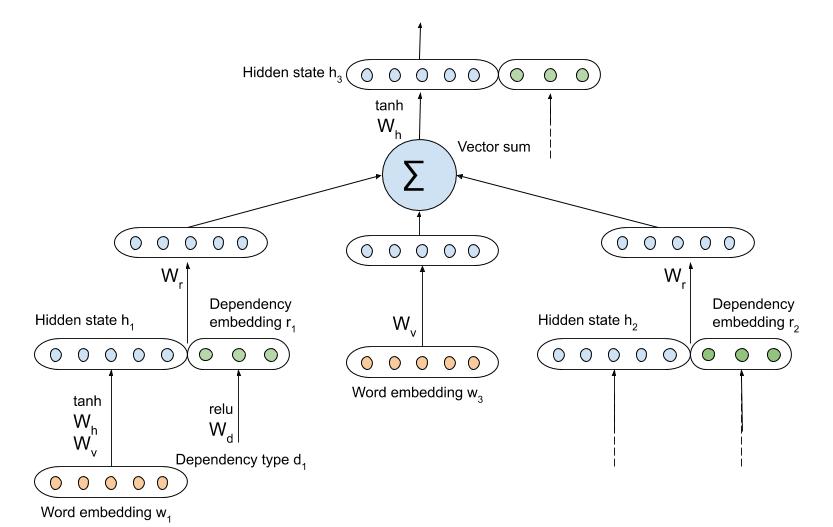}
  \caption{Semantic composition using the proposed Typed Dependency Tree-RNN for \emph{node $3$} with two children \emph{node $1$} and \emph{ node $2$} related by dependency type $r_1$ and $r_2$ repectively. $W_v,W_h,W_r$ and $W_d$ are the set of shared weights of the model. \emph{Node $1$} is a leaf node.}
  \label{fig:tydtrnn}
\end{figure}
Starting from the leaf nodes, the model computes a vector representation $h_t$ of each node $t$. The hidden state $h_t$ encodes the meaning of the sub-tree rooted at $t$.
At every node t, the Typed DT-RNN unit takes as inputs -- $x_t$ the word vector of the current word $w_t$ and a list of $m$ tuples $(h_k,z_{k})$ where $h_k$ is hidden state of the child node $k$, and $z_{k}$ the one-hot encoding of relation between the words at nodes $k$ and $t$. Equation \ref{eqn:tdtrnn} denotes the set of transitions at node $t$ with $m$ child nodes. The typed dependency of each node pair $(k,t)$ is embedded into a real valued vector $d_k$, which is than concatenated with the hidden state $h_k$. The hidden state of the node $t$ is a function of $x_t$ and the concatenated vector $h_k \, :\,d_k$ where $k \, \epsilon \, C(t)$ - set of child nodes of $t$, as illustrated in Fig. \ref{fig:tydtrnn}. The transition function for computing hidden state of node $t$ is given by equation \eqref{eqn:tdtrnn}.

\begin{align}
d_k &= g ( W_d \;z_k + b_{d}),\nonumber\\
h_t &= f \left( \frac{1}{l(t)} \left(W_v x_t + \sum_{k \epsilon C(t)} l(k) W_{r} \left(h_k:d_k \right) \right) \right),\label{eqn:tdtrnn}\\ 
l(t) &= 1 + \sum_{k \epsilon C(t)} l(k) \nonumber
\end{align}
where $g$ and $f$ are activation functions. We found ReLU (Rectified Linear Unit) to be optimal for $g$ and $tanh$ is used as function $f$.

Unlike the DT-RNN/SDT-RNN, which uses a different weight matrix for each position/dependency type, our model uses the same shared weight $W_r\,\epsilon\, R^{(h+r)\times h}$ for every child node. The typed dependency information is incorporated by concatenating the dependency embedding $d_k$ with each child node's hidden state based on the relationship child $k$ shares with its parent. We have eliminated the need for separate dependency matrices using dependency vectors. The proposed model has lesser parameters to learn, thereby is lesser prone to overfitting.

\section{Experimental setup}\label{sec:exp}
This section describes the datasets and the training details of the experiments.
\subsection{Datasets}
The experiments use the SICK \cite{marelliraffaella} dataset designed to test the compositional ability of models. Compared to other sentential datasets, SICK is rich in the lexical, syntactic, and semantic phenomena for which the compositional models are expected to account. The dataset contains 9927 sentence pairs annotated with an inference tag (Contradiction/Neutral/Entailment) and a relatedness score ranging from 1 (least related) to 5 (the most related). The SICK dataset's sentence pairs are sentences extracted from the 8K ImageFlickr dataset and the SemEval 2012 STS MSR-Video Description data set. The relatedness score is the average of manually assigned ten scores.  The SICK dataset contains separate subsets for training (4500), development (500), and testing (4927). 
\subsection{Hyperparameter \& Training details}
The pre-trained Glove \cite{pennington2014glove} vectors of 300-dimension are used as fixed word embedding. All other parameters, randomly initialized, are trainable. We generate the parse trees using Stanford dependency parser\cite{chen2014fast,manning-EtAl:2014:P14-5}, which identifies 47 types of universal dependencies. The model uses one-hot encodings to input these dependency types. The model is trained end-to-end by a mini-batch gradient descent algorithm using AdaGrad optimizer. The Table \ref{tab:hyperp} lists the optimal hyperparameters chosen based on the performance on the development dataset.
\begin{table}  
\centering
\caption{Hyperparameter settings}
\begin{threeparttable}
\begin{tabular}{@{}p{5cm}p{1.5cm}p{1.5cm}@{}}
\hline \hline
Parameters & SICK-R & SICK-E \\\hline
Learning rate  &0.01 & 0.015\\
Batch size & 25 & 10\\
Classifier's hidden layer size (c)& 100  & 100\\
RNNs Hidden state dimension (h) & 130 & 100\\
Typed dependency embedding size (r) & 30 & 10\\
Number of epochs & 14 & 26\\
Weight decay for Adagrad & 0.0001 & 0.0001\\
\hline
\end{tabular} 
\label{tab:hyperp} 
\end{threeparttable}
\end{table}
\section{Results and Discussion}\label{sec:results}
Table \ref{tab:resultsent} lists the comparison of the accuracy of the proposed model with other state-of-the-art Tree-RNN models for entailment classification. The models are grouped as: (1) Recursive NN variants, (2) Sequential LSTM variants, and (3) TreeLSTM variants. Pearson's correlation, Spearman's correlation, and Mean Squared Error between the predicted relatedness score and human rating for the SICK test dataset are listed in Table \ref{tab:resultssrs}. 
The DT-RNN-single is our implementation of DT-RNN using a single shared matrix instead of separate location-based weights. The results show that the proposed Typed DT-RNN model with fewer parameters outperforms the other variants of recursive NN models in both scoring and classification tasks.  Except for the Bi-LSTM model, the classification accuracy of typed DT-RNN is comparable with that of sequential LSTMs. For both tasks, TreeLSTMs outperform Tree-RNN as LSTMs have memory states that can retain information over longer sequences.

Table \ref{tab:exsent} lists some of the sentence pairs from the SICK test dataset along with the actual inference label and the label assigned by the proposed model. The output shows that the proposed model is more efficient in identifying entailments and contradictions caused by words in significant roles like the subject, object, and verbs. The model misclassified pairs in which adjective modifiers caused a contradiction in meaning, e.g.: ``one leg'' and ``two legs''. A detailed investigation of the learned typed embedding and its effect on classification errors can shed more light on how neural networks understand sentence semantics.
\begin{figure*}[!t]
\centering
\subfloat[Mean Squared Error (MSE)]{\includegraphics[width=3in]{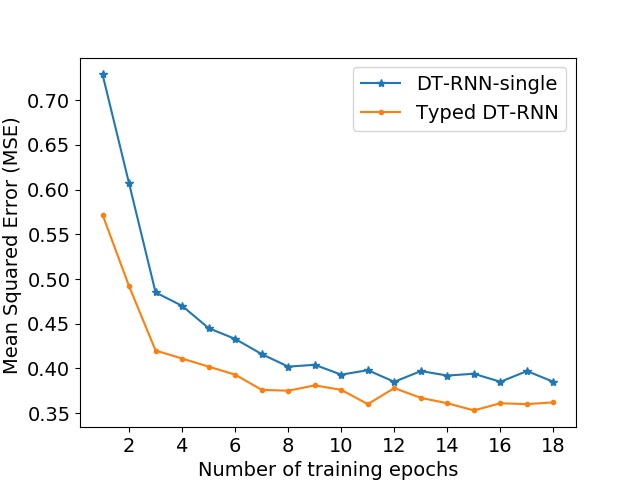}}
\hfil
\subfloat[Pearson correlation coefficient ( $r \times 100$)]{\includegraphics[width=3in]{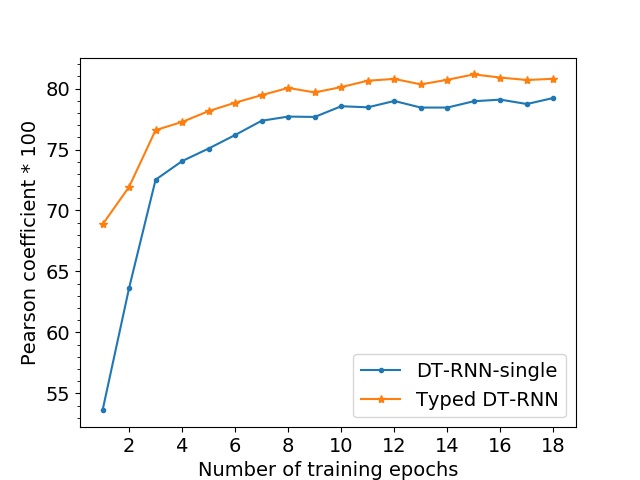}}
\caption{Comparison of learning curves of the proposed Typed DT-RNN and DT-RNN (using a single shared weight for all dependencies) using SICK test dataset.}
\label{fig:curves}
\end{figure*}
\begin{table}
\centering
\caption{Classification accuracies on SICK-E test set.}
\begin{threeparttable}
\begin{tabular}{@{}p{5cm}p{2cm}@{}}
\hline \hline
Model & Accuracy (\%) \\\hline
Recursive NN\cite{socher2011parsing} & 74.9\\
DC-RecNN \cite{kim2019dynamic} & 77.9\\
RNTN\cite{socher2013recursive} & 76.9 \\
MV-RNN\cite{socher2012semantic} & 75.5\\
DT-RNN\cite{socher2014tacl}  & 63.38\\
SDT-RNN\cite{socher2014tacl}$^\star$  & 78.30\\
DT-RNN-single$^\star$ & 78.10\\
\hline
LSTM\cite{tai2015improved} & 76.80\\
Bi--LSTM\cite{tai2015improved} & 82.11\\
2-layer LSTM\cite{tai2015improved} & 78.54\\
2-layer Bi-LSTM\cite{tai2015improved} & 79.66\\\hline
DT-LSTM\cite{tai2015improved} & 83.11\\
CT-LSTM\cite{tai2015improved} & 82.00\\
DC-TreeLSTM \cite{kim2019dynamic} & 80.2\\
\hline
\textbf{Typed DT-RNN} (proposed) & \textbf{80.54} \\\hline
\end{tabular}
\begin{tablenotes}
 \item [$\star$] Our implementation.
\end{tablenotes}
\end{threeparttable}
\label{tab:resultsent}
\end{table}
\begin{table}
\centering
\caption{Comparison of Typed Dependency Tree-RNN with other RNN models for semantic relatedness scoring.}
\begin{tabular}{@{}p{2.5cm}p{1.5cm}p{2cm}p{1cm}@{}}
\hline \hline
Model & Pearson's $r$ & Spearman's $\rho$& MSE \\\hline
Mean vectors & 0.7577 & 0.6738 & 0.4557\\
\hline
D-LSTM & 0.8270 & 0.7673 & 0.3527\\
pos-LSTM-all & 0.8173 & 0.7610 &-- \\
\hline
Typed DT-LSTM & 0.8731 &--& 0.2427\\
DT-LSTM & 0.8676 & 0.7729 & 0.2532\\
\hline
DT-RNN & 0.7923 & 0.7319 & 0.3848\\
SDT-RNN & 0.7900 & 0.7304 & 0.3848\\
DT-RNN-single & 0.7966 & 0.7334 & 0.3848\\\hline
\textbf{Typed DT-RNN} & \textbf{0.8116} & \textbf{0.7426}&\textbf{0.353}\\\hline
\end{tabular}
\label{tab:resultssrs}
\end{table}
\begin{table} 
\caption{Sample sentence pairs from SICK test dataset with predicted labels P and actual labels G.}
\centering
\begin{threeparttable}
\begin{tabular}{@{}p{6cm}p{1cm}p{1cm}@{}}
\hline \hline
Sentence pairs & P & G\\\hline
\hline
A group of scouts are hiking through the grass. & & \\ People are walking. & E & E\\
\hline
A white and brown dog is walking through the water with difficulty & &\\A dog that has a brown and white coat is trotting through shallow water & E & N\\\hline
Nobody is pouring ingredients into a pot & &\\	Someone is pouring ingredients into a pot. & C & C \\
\hline
A child is hitting a baseball & & \\	A child is missing a baseball & C & C \\\hline
A young girl is standing on one leg	& & \\ A young girl is standing on both legs & E & C\\\hline
The dog is running through the wet sand. & &\\ The dog is running through the dry sand & E & C \\\hline
A brown dog is attacking another animal in front of the man in pants & & \\There is no dog wrestling and hugging & C & N\\
\hline
\end{tabular}
\label{tab:exsent}
\begin{tablenotes}
    \item C -- Contradiction, N -- Neutral, and E -- Entailment
  \end{tablenotes}
\end{threeparttable}
 \end{table}

The learning curves of the proposed typed DT-RNN along with DT-RNN-single are plotted in Fig. \ref{fig:curves}. We find that the proposed model outperforms the DT-RNN-single in a lesser number of epochs. 

\section{Conclusion and Futureworks}\label{sec:Conclusion}
The primary contribution of this paper is a novel Typed DT-RNN model for composing the sentence vectors from its constituent word vectors and the grammatical dependency relations between them. Our experiments show that the model is superior to DT-RNN and SDT-RNN in semantic similarity scoring and recognizing textual entailment tasks. The paper also presents a comparison of the results with other Tree-RNN models. 
Though many works in literature have attempted to incorporate grammatic roles like POS tags and role labels into deep learning models, the role of typed dependencies has not yet been much explored. A detailed qualitative analysis of learned typed embeddings can give better insight into language understanding.  As future work, we propose to extend this work to other semantic matching tasks like paraphrase identification and question answering. 
%
%

\end{document}